\documentclass[runningheads, envcountsame,a4paper]{llncs}
\usepackage{color,soul}
\usepackage{graphicx}
\usepackage{romannum}
\usepackage{balance}
\usepackage{subfig}
\usepackage{lipsum}
\usepackage{answers}
\usepackage{setspace}
\usepackage{graphicx}
\usepackage{enumitem}
\usepackage{multicol}
\usepackage{mathrsfs}
\usepackage[utf8]{inputenc}
\usepackage[misc]{ifsym}
\newcommand{\x}{\mathbf{x}}
\newcommand{\uu}{\mathbf{u}}

\newcommand{\bb}{\mathbf{b}}
\newcommand{\w}{\mathbf{w}}
\newcommand{\W}{\mathbf{W}}
\newcommand{\CW}{\mathcal{W}}
\newcommand{\E}{\mathrm{E}}
\newcommand{\A}{\mathcal{A}}
\newcommand{\OO}{\mathcal{O}}
\newcommand{\B}{\mathcal{B}}

\newcommand{\LL}{\mathcal{L}}
\newcommand{\N}{\mathcal{N}}
\newcommand{\M}{\mathcal{M}}
\newcommand{\UU}{\mathcal{U}}
\newcommand{\ANLP}{{\it ANLP }}

\usepackage{amsmath,amssymb}

\begin{document}
    \title{Scalable Bid Landscape Forecasting in Real-time Bidding}
    %
    %
    \author{Aritra Ghosh\thanks{This work was conducted while the first author was doing an internship at
            Adobe Research, USA  }\inst{1} (\Letter) \and
        Saayan Mitra\inst{2} \and
        Somdeb Sarkhel\inst{2} \and Jason Xie\inst{3} \and Gang Wu\inst{2} \and Viswanathan Swaminathan \inst{2}}
    \authorrunning{A. Ghosh et al.}
    %
    \institute{University of Massachusetts, Amherst, MA, USA \and
        Adobe Research, San Jose, CA, USA \and Adobe Advertising Cloud, Emeryville, CA, USA\\
        \email{arighosh@cs.umass.edu,}
        \email{$\{$smitra,sarkhel,jasonxie, gawu,vishy$\}$@adobe.com}}
    \maketitle              

    \begin{abstract}
        In programmatic advertising, ad slots are usually sold using second-price (SP) auctions in real-time. The highest bidding advertiser wins but pays only the second highest bid (known as the {\it winning price}). In SP, for a single item, the dominant strategy of each bidder is to bid the true value from the bidder's perspective. However, in a practical setting, with budget constraints, bidding the true value is a sub-optimal strategy. Hence, to devise an optimal bidding strategy, it is of utmost importance to learn the winning price distribution accurately. Moreover, a demand-side platform (DSP), which bids on behalf of advertisers, observes the winning price if it wins the auction. For losing auctions, DSPs can only treat its bidding price as the lower bound for the unknown winning price. In literature, typically censored regression is used to model such partially observed data. A common assumption in censored regression is that the winning price is drawn from a fixed variance (homoscedastic) uni-modal distribution (most often Gaussian). However, in reality, these assumptions are often violated.
        We relax these assumptions and propose a heteroscedastic fully parametric censored regression approach, as well as a mixture density censored network.
        Our approach not only generalizes censored regression but also provides flexibility to model arbitrarily distributed real-world data. Experimental evaluation on the publicly available dataset for winning price estimation demonstrates the effectiveness of our method. 
        Furthermore, we evaluate our algorithm on one of the
        largest demand-side platform and significant improvement has been achieved
        in comparison with the baseline solutions.
    \end{abstract}
    \section{Introduction}
    Real-time Bidding (RTB) has become the dominant mechanism to sell ad slots over the internet in recent times.
    In RTB, ad display opportunities are auctioned when available from the publishers (sellers) to the advertisers (buyers). When a user sees the ad that won the auction, it is counted as an {\it ad impression}.  
    An RTB ecosystem consists of supply-side platforms (SSP), demand-side platforms (DSP) and an Ad Exchange.  When a user visits a publisher's page, the SSP sends a request to the Ad Exchange for an ad display opportunity which is then rerouted to DSPs in the form of a bid request. DSPs bid on behalf of the advertisers at the Ad Exchange. The winner of the auction places the Ad on the publisher's site. Ad Exchanges usually employ second-price auction (SP) where the winning DSP only has to pay the second highest bidding price \cite{yuan2013real}. Since this price is the minimum bidding price DSP needs to win, it is known as the {\it winning price}. When a DSP wins the auction, it knows the actual winning price. However, if the DSP loses the auction, the Ad Exchange does not reveal the winning price. In that case, the bidding price provides a lower bound on the winning price. This mixture of observed and partially-observed (lower bound) data is known as {\it right censored data}. The data to the {\it right} of the bidding price is not observed since it is right censored.
    
    For a single {\it ad impression} under the second price auction scheme, the dominant strategy for an advertiser is to bid the true value of the ad. In this scenario, knowing the bidding prices of other DSPs does not change a bidder's strategy \cite{edelman2007internet}. However, in reality, DSPs have budget constraints with a utility goal (e.g., number of impressions, clicks, conversions). Under budget constraints, with repeated auctions, bidding the true value is no longer the dominant strategy \cite{balseiro2015repeated}. In this setting, knowledge of the bidding prices of other bidders can allow one to change the bid to improve its expected utility. DSP needs to estimate the cost and utility of an auction to compute the optimal bidding strategy (or bidding price)  \cite{zhang2014optimal}. To compute the expected cost as well as the expected utility one needs to know the winning price distribution.
    Therefore, modeling the winning price distribution is an important problem for a DSP \cite{lang2012handling}. This problem is also referred to as the {\it Bid landscape forecasting} problem.
    
    Learning the bid landscape from a mix of observed and partially-observed data poses a real challenge. It is not possible for DSPs to know the behavior of the winning price beyond the maximum bidding price. Parametric approaches often assume that the winning price follows some distribution. In the existing literature, Gaussian and Log-Normal distributions are often used for modeling the winning price \cite{wu2015predicting,cui2011bid}. 
    However, these simple distributions do not always capture all the complexities of real-world data. 
    Moreover, for losing bids, the density of winning price cannot be measured directly, and hence a standard log-likelihood based estimate does not typically work on the censored data. In this scenario, a common parametric method used is {\it Censored Regression}, which combines the log density and the log probability for winning and losing auctions respectively \cite{wu2015predicting,powell1984least}. Another common alternative is to use non-parametric survival based methods using the Kaplan-Meier (KM) estimate for censored data \cite{kaplan1958nonparametric}. To improve the performance of the KM estimate, clustering the input is important.
    Interestingly, in {\cite{wang2016functional}}, the authors proposed to grow a decision tree based on survival methods. In the absence of distributional assumptions, non-parametric methods (KM) work well. 
However, efficiently scaling non-parametric methods is also challenging. On the other hand, parametric methods work on strong distributional assumptions. When the assumptions are violated, inconsistency arises. For a general discussion of the censored problem in machine learning, readers are referred to {\cite{wang2017machine}}.
    
    Learning a distribution is generally more challenging than point estimation. 
    Thus,    parametric approaches in previous research often considered point estimation \cite{wu2015predicting,wu2018deep}. However, to obtain an optimal bidding strategy, one needs the distribution of the winning price. On the other hand, non-parametric approaches like the KM method computes the distribution without any assumptions.
    However, these methods require clustering the data to improve the accuracy of the model using some heuristics.
    Clustering based on feature attributes makes these methods sub-optimal impacting generalization ability for dynamic real-world ad data. 

\begin{figure}%
        \centering
        \subfloat[Cluster 1 ]{{\includegraphics[width=6.1cm]{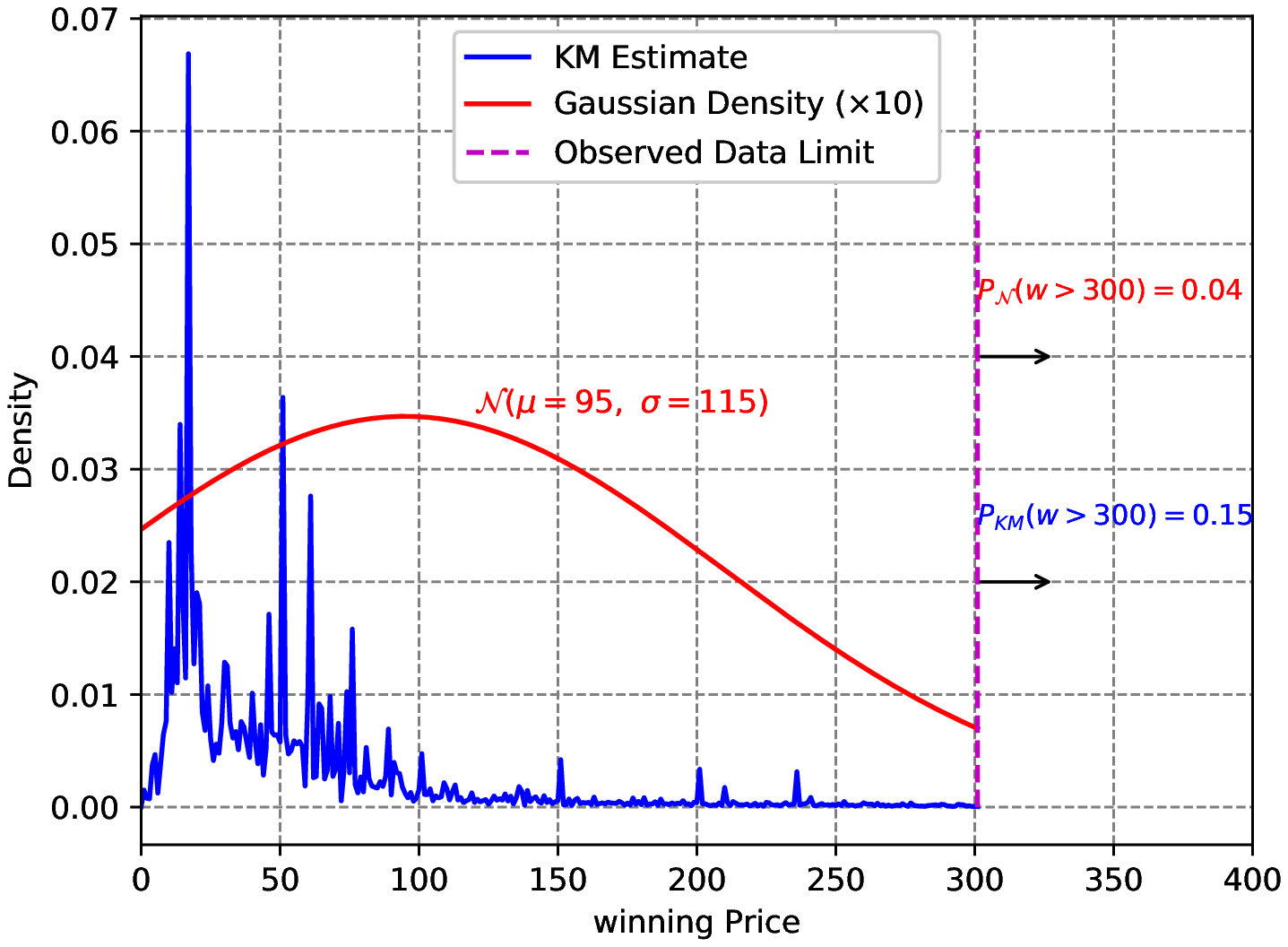} }}%
        \subfloat[Cluster 2 ]{{\includegraphics[width=6.1cm]{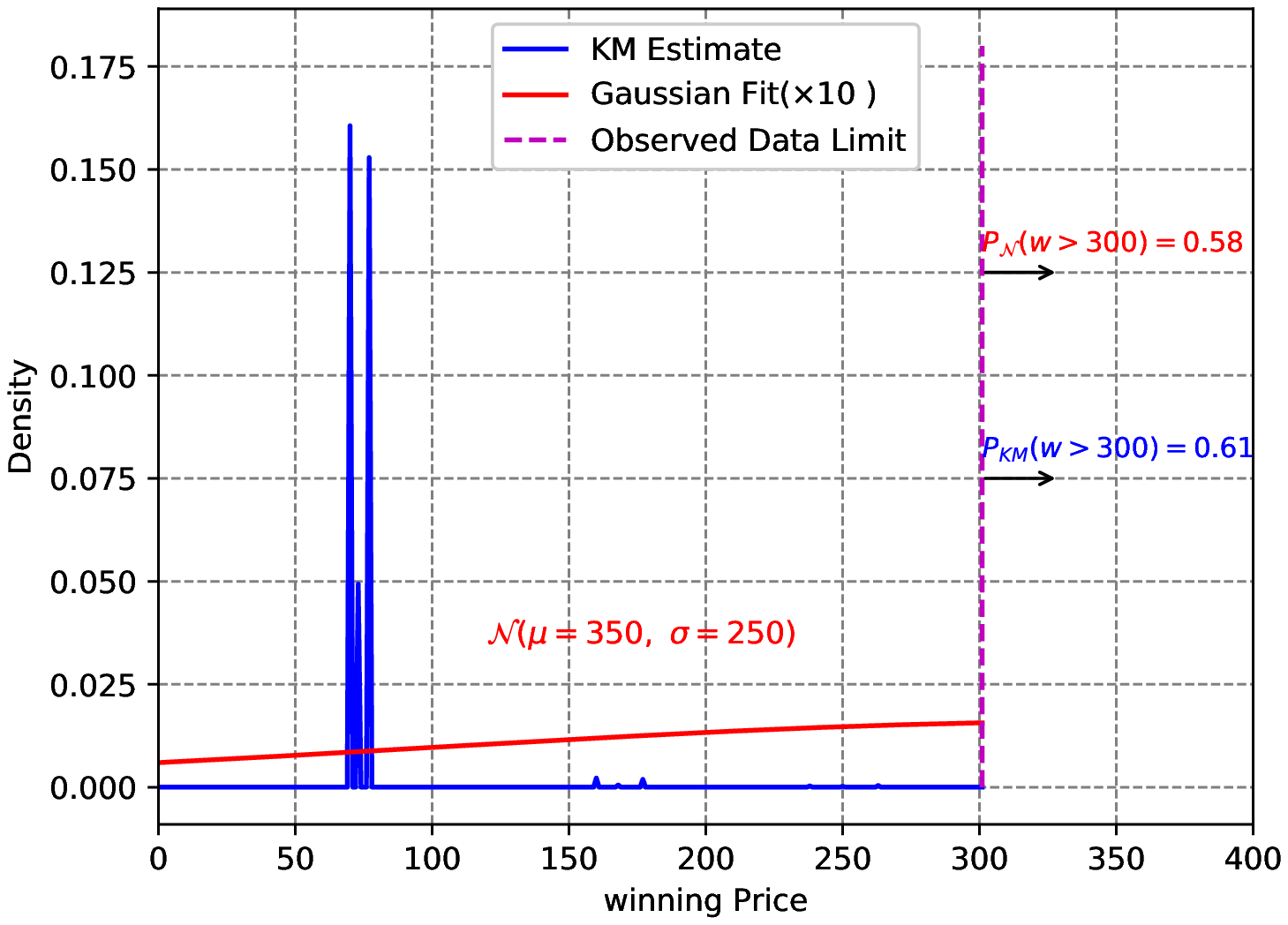} }}%
        \caption{KM Estimate and Gaussian Fit on two clusters of Session-2 date 2013-06-12 on iPinYou \cite{zhang2014real}}
        \label{fig:km-estimate}%
    \end{figure}

    In this paper, we close the gap of violated assumptions in parametric approaches on censored data.
    Censored regression-based approaches assume a unimodal (often Gaussian) distribution on winning price.
    Additionally, it assumes that the standard deviation of the Gaussian distribution is unknown but fixed.
    However, in most real-world datasets these assumptions are often violated. For example, in Figure~\ref{fig:km-estimate}, we present two winning price distributions (learned using the KM estimate) as well as fitted Gaussian distributions\footnote{We fit the unimodal Gaussian minimizing KL divergence with the estimated KM distribution. We would like to point out, although in Figure~\ref{fig:km-estimate}(b) winning price density is unimodal (within the limit), the probability of winning price beyond the max bid price is high ($0.61$). Thus the fitted Gaussian has a mean of $350$ and std dev of $250$ further from the peak at $75$.} on two different partitions of the iPinYou dataset \cite{zhang2014real}. It is evident from Figure~\ref{fig:km-estimate} that the distributions are neither   Gaussian (blue line) nor have fixed variance (red line). In this paper, we relax each of these assumptions one by one and propose a general framework to solve the problem of predicting the winning price distribution using partially observed censored data. We first propose an additional parameterization which addresses the fixed variance assumption.
    Further, the Mixture Density Network is known to approximate any continuous, differentiable function with enough hidden nodes {\cite{bishop1994mixture}}. 
    We propose a Mixture Density Censored Network to learn smooth winning price distribution using the censored data. We refer to it as MCNet in the rest of the paper.  Both of our proposed approaches are generalizations of the Censored Regression. 
    
    Our main contributions are as follows. The typical deployed system uses Censored regression for point estimation of the winning price. However, we argue that point estimation is not enough for an optimal bidding strategy. 
    We improve upon the parametric Censored Regression model to a general framework under minimal assumptions. We pose Censored Regression as a solution to the winning price distribution estimation problem (instead of a point estimate). To the best of our knowledge, we are the first to apply the mixture density network on censored data for learning the arbitrary distribution of the winning price. Our extensive experiments on a real-world public dataset show that our approach vastly out-performs existing state-of-the-art approaches such as \cite{wu2015predicting}. Evaluation on the historical bid data from Adobe (DSP) shows the efficacy of our scalable solution. While we restricted the analysis to winning price distribution in real-time bidding, MCNet is applicable to any partially observed censored data problem.

    \section{Background \& Related Work}
    \label{sec:back}
    In RTB, a DSP gets bid requests from the Ad exchange. We represent the $i^{th}$ bid request by a feature vector $\x_i$, which captures all the characteristics of the bid request. Most of the elements of $\x_i$'s are categorical (publisher verticals, user's device, etc.). If DSP wins the auction, it pays the second (winning) price. Formally, the winning price is, 
    \[\w_i= \max\{\bb^{\text{Pub}}_i, \bb^{\text{DSP}_1}_i, \bb^{\text{DSP}_2}_i, \cdots, \bb^{\text{DSP}_K}_i\}\]
    where $\bb^{\text{Pub}}_i$ is the floor price set by the publisher\footnote{For simplicity, we view the floor price by the publisher as a bid from an additional DSP.} (often 0), and $\bb^{\text{DSP}_1}_i, \cdots, \bb^{\text{DSP}_K}_i$ are bidding prices from all other participating DSPs. We use $\bb_i$ to denote the bidding price from the DSP of our interest.
    Here we provide an example to illustrate the winning price (in SP auction). Suppose DSPs A, B, C bid $\$1$, $\$2$, $\$3$ respectively for a bid request. DSP C then wins the auction and pays the second-highest price, i.e., $\$2$. For DSP C, the winning price is $\$2$ (observed). For losing DSPs, A, and B, the winning price is $\$3$ (which is unknown to them). In this paper, we define the winning price from the perspective of a single DSP.
    
    Learning the landscape of winning price accurately is important for an optimal bidding strategy.
    A DSP is usually interested in some utility $\uu_i$ (e.g., clicks, impressions, conversions)  for each bid request $\x_i$ and wants to maximize the overall utility using bidding strategy $\A$ and with budget $\B$. This can be represented by the following optimization problem,
    $\max_{\A} \sum_{i} \uu_i \mbox{ s.t. } \sum_{i} cost_i\leq \B$,
    where $cost_i$ is the price the DSP pays, if it wins the auction. Although the variables are unknown beforehand,
    the expected cost and the utility can be computed using the historical bid information. Thus the problem simplifies to,
    \begin{align}
    \max_{\A} &\sum_{i}\E [\uu_i|\x_i, \bb_i] \mbox{ s.t. } \sum_{i} \E [ cost_i|\x_i, \bb_i]\leq \B
    \end{align}
    Note that, the expected utility $\uu_i$ is conditioned on  bid request $\x_i$ and  the actual bid $\bb_i$.
    For bid request $\x_i$, we represent the winning price distribution as $P_{\w}(\W_i|\x_i)$, and its cumulative distribution function (cdf) as $F_{\w}(\W_i|\x_i) $. If the DSP bids $\bb_i$ for $\x_i$, 
    expected cost and expected utility (for SP auction) is, \[\E [cost_i|\x_i, \bb_i] = {\int_0^{\bb_i} \w P_{\w}(\W_i=\w|\x_i)d\w} ,\quad  \ \E[\uu_i|\x_i,\bb_i]= F_{\w}(\bb_i|\x_i) \E[\uu_i|\x_i]\]
    An example of expected utility conditioned on bid request ($\E[\uu_i|\x_i]$) is Click-through rate (CTR). CTR prediction is a well-studied problem in academia and the industry \cite{wang2017deep}.
    We want to point out that the expected cost ($\E [cost_i|\x_i, \bb_i] $) is not the same as the expected winning price ($\E[\W_i|\x_i]$). The former is always lower than the latter and is equal only when $\bb_i\rightarrow \infty$ (i.e., when the advertiser wins the auction with probability 1 and observe the winning price). 
    Thus predicting the winning price distribution instead of the point estimate is important \cite{wang2016display}. Further, for pacing the budget, one requires an estimate of winning price distribution \cite{agarwal2014budget}. In \cite{zhang2016bid}, the authors proposed an unbiased learning algorithm of click-through rate estimation using the winning price distribution.
    Earlier parametric methods, considered point estimation of the winning price. 
    The censored regression-based approach assumes a standard unimodal distribution with a fixed but unknown variance to model the winning price \cite{wu2015predicting,zhu2017gamma,wang2017deep}. 
    In another paradigm, non-parametric methods such as the KM estimator has been successful for modeling censored data \cite{kaplan1958nonparametric,wang2016functional}. 
    
    In the rest of the paper, we use $P$ to denote probability density function (pdf) and $\Pr$ to denote the usual probabilities. 
    Next, we describe how Censored Regression is applied to model the winning price.
    
    \subsection{Censored Regression}
    The data available to DSP is right censored by the Ad Exchange, i.e., for losing bids only a lower bound (the bidding price) of the winning price is known. However, a maximum likelihood estimator (MLE) can still work on the censored data with some assumptions. 
    
    In \cite{wu2015predicting}, the authors assume that the winning price follows a normal distribution with fixed but unknown variance $\sigma$. The authors assume a linear relationship between the mean of the normal distribution and the input feature vector. 
    We use $\W_i$ to represent the random variable of winning price distribution  of $i^{\mbox{th}}$ bid request whereas $\w_i$ is the realization of that. Thus $\w_i = \beta^T \x_i + \epsilon_i$ where $\epsilon_i$  are independent and identically distributed ({\it i.i.d}) from $\N(0, \sigma^2)$ and $\W_i \sim \N (\beta^T\x_i, \sigma^2)$.
    
    One can use any standard distribution in the censored regression approach. 
    In {\cite{wu2018deep}}, the authors argue that maximal bidding price in the limit (of infinite DSPs) resembles Gumbel distribution. However, for the generality of learning from censored data, we do not constrain on any particular distribution in this paper.
    Moreover, the linear link function can be replaced with any non-linear function. Thus, $\w_i$ can be parameterized as $\w_i= f(\beta, \x_i)+\epsilon_i$ where $f$ can be any continuous differentiable function. With the success of deep models, in {\cite{wu2018deep}}, the authors parameterize $f(\beta, \x_i)$  with a deep network for additional flexibility. 
    Since we know the winning price for winning auctions, likelihood is simply the probability density function (pdf) $P(\W_i = \w_i) =   \frac{1}{\sigma}\phi(\frac{\w_i - \beta^T\x_i}{\sigma})$ where $\phi$ is the pdf of standard normal $\N(0,1)$. Note that, $\W_i$ is the random variable associated with the winning price distribution whereas $\w_i$ is the observed winning price.
    For losing auctions, as we do not know the winning price, the pdf is unknown to us. However, from the lower bound on the winning price, we can compute the probability that bid $\bb_i$ will lose in the auction for bid request $\x_i$, under the estimated distribution of $\W_i$ as $    \Pr(\W_i>\bb_i) = \Pr(\epsilon_i < \beta^T\x_i -\bb_i   ) = \Phi(\frac{\beta^T\x_i - \bb_i}{\sigma}).$
    Here $\Phi$ is the cdf for standard normal distribution. As discussed, $\phi$ and $\Phi$ can be replaced with pdf and cdf of any other distribution (with different parameterization).

    Taking log of the density for winning auctions $\CW$ and the log-probability for losing auctions $\LL$, we get the following objective function \cite{wu2015predicting},
    \begin{align}
    \beta^{\ast}, \sigma^{\ast} =& \mbox{arg}\max_{\beta, \sigma>0} \sum_{i \in \CW} {\log\left(\frac{1}{\sigma} \phi(\frac{\w_i - \beta^T\x_i}{\sigma}) \right)}
    + \sum_{i \in \LL} {\log\left( \Phi(\frac{\beta^T\x_i - \bb_i}{\sigma}) \right)}
    \label{eq:censored}
    \end{align}
    When the $\epsilon_i$ (noise in the winning price model) are i.i.d samples from a fixed variance normal distribution, censored regression is an unbiased and consistent estimator \cite{james1984consistency,greene1981asymptotic}.\\
    \section{Methodology}
    
    In this paper, we build on top of (Gaussian) censored regression-based approach by relaxing some of the assumptions that do not hold in practice. First, we relax the assumption of \textit{homoscedasticity}, i.e., noise (or error) follows a normal distribution with fixed but possibly unknown variance, by modeling it as a fully parametric censored regression. Then we also relax the unimodality assumption by proposing a mixture density censored network. We describe the details of our approaches in the next two subsections.
    
    \subsection{Fully Parametric Censored Regression}
    The censored regression approach assumes that the winning price is normally distributed with a fixed standard deviation. As we discussed, in Figure~\ref{fig:km-estimate}, the variance of the fitted Gaussian model is not fixed. If the noise $\epsilon$ is heteroscedastic or not from a fixed variance normal distribution, the MLE is biased and inconsistent. Using a single $\sigma$ to model all bid requests, will not fully utilize the predictive power of the censored regression model. Moreover, while the point estimate (mean) of the winning price is not dependent on the estimated variance, the {\it Bid landscape} changes with $\sigma$. We remove the restriction of  homoscedasticity in the censored regression model and pose it as a solution to the distribution learning problem. 
    
    Specifically, we assume the error term $\epsilon$ is coming from a parametric distribution conditioned on the features.  This solves the problem of error term coming from fixed variance distribution. We assume the noise term  $\epsilon_i$ is coming from $\N(0, \sigma_i^2)$ where $\sigma_i = \exp(\alpha^T\x_i)$. 
    
    The likelihood for winning the auction is, $P(\W_i =\w_i) =   \frac{1}{\exp(\alpha^T\x_i)}\phi(\frac{\w_i - \beta^T\x_i}{\exp(\alpha^T\x_i)})$ where  the predicted random variable $\W_i \sim \N(\beta^T\x_i , \exp(\alpha^T\x_i)^2)$ and $\phi$ is the pdf of $\N(0,1)$.
    In fully parametric censored regression, $\epsilon_i\sim \N (0, \exp(\alpha^T\x_i)^2)$ are not {\it i.i.d} samples. For losing bids, we can similarly compute the probability based on the lower bound (bidding price $\bb_i$)
    \begin{align*}
    \Pr(\W_i >\bb_i) = P(\epsilon_i < \beta^T\x_i -\bb_i   ) = \Phi(\frac{\beta^T\x_i - \bb_i}{\exp(\alpha^T\x_i)})
    \end{align*}
    
    Under the assumption of normal but varying variance on the noise, we can still get a consistent and unbiased estimator by solving the following problem. 
    
    \begin{align}
    \beta^{\ast}, \alpha^{\ast} =& \mbox{arg}\max_{\beta, \alpha} \sum_{i \in \CW} {\log\left(\frac{1}{\exp(\alpha^T\x_i)} \phi(\frac{\w_i - \beta^T\x_i}{\exp(\alpha^T\x_i)}) \right)}
    +\sum_{i \in \LL} {\log\left( \Phi(\frac{\beta^T\x_i - \bb_i}{\exp(\alpha^T\x_i)}) \right)}
    \label{eq:parm_censored}
    \end{align}
    

    \subsection{Mixture Density Censored Network (MCNet)}
    In the previous subsection, we relaxed the fixed variance problem by using a parametric $\sigma$. However, no standard distribution can model the multi-modality that we observe in real-world data. For example, in Figure~\ref{fig:km-estimate}(b), we see mostly unimodal behavior below the max bid price. However, the probability of losing an auction is often high ($61\%$ in Figure ~\ref{fig:km-estimate}(b)). Thus even with parametric standard deviation, when we minimize the KL-divergence with a Gaussian, the mean shifts towards the middle. Inspired by the Gaussian Mixture Model (GMM)~\cite{bishop1994mixture} we propose a Mixture Density Censored Network (MCNet). MCNet resembles a Mixture Density Network while handling partially observed censored data for learning arbitrary continuous distribution.
    
    In a GMM, the estimated  random variable $\W_i$ consists of $K$ Gaussian densities and has the following pdf, $P(\W_i =\w_i) = \sum_{k=1}^{K} \frac{\pi_k(\x_i)}{ \sigma_k(\x_i)} \phi(\frac{\w_i-\mu_k(\x_i)}{ \sigma_k^2(\x_i)})$.
    Here $\pi_k(\x), \mu_k(\x), \sigma_k(\x)$ are the weight, mean and standard deviation for $k^{th}$ mixture density respectively where $k \in \{1, \cdots, K\}$.
    To model the censored regression problem as a mixture model, a straightforward way is to formulate the mean of the Gaussian distributions with a linear function. Furthermore, to impose positivity of $\sigma$, we model the logarithm of the standard deviation as a linear function. We impose a similar positivity constraint on the weight parameters. The parameters of the mixture model are (for $k\in \{1, \cdots, K\}$),
    \begin{align*}
    \mu_k(\x_i) = \beta_{\mu, k}^T \x_i, \
    \sigma_k(\x_i) = \exp(\beta_{\sigma, k}^T\x_i),\ 
    \pi_k(\x_i) = \frac{\exp(\beta_{\pi, k}^T\x_i)}{\sum_{j=1}^{K} \exp(\beta_{\pi,j}^T\x_i)}
    \end{align*}
    We can further generalize this mixture model and define a Mixture Density Network (MDN) by parameterizing $\pi_k(\x_i), \mu_k(\x_i), \sigma_k(\x_i)$ with a deep network. In applications such as speech and image processing and astrophysics, MDNs have been found useful \cite{zen2014deep,salimans2017pixelcnn++}. MDN can work with any reasonable choice of base distribution. 
    
    MDN combines mixture models with neural networks. The output activation layer, consists of $3K$ nodes ($z_{i,k}$ for $i \in \{\mu, \sigma,\pi\}$ and $k\in \{1, \cdots, K\}$      ).  We use $z_{\mu,k}, z_{\sigma,k}, z_{\pi,k}$ to retrieve the mean, standard deviation and weight  parameters of $k^{th}$ density,
    \begin{align}
    \mu_k(\x_i) = z_{\mu,k}(\x),\  
    \sigma_k(\x_i)  = \exp (z_{\sigma,k } (\x)),\  
    \pi_k(\x_i) = \frac{\exp(z_{\pi, k}(\x_i))}{ \sum_{j=1}^{K }\exp(z_{\pi,j} (\x_i))                }\label{eq:pi}
    \end{align}
    
    MDN outputs conditional probabilities that are used for learning distribution from fully observed data {\cite{bishop1994mixture}}. For the censored problem, however, we only observe partial data. We can now extend MDN to MCNet on censored data. Instead of conditional output probabilities, MCNet outputs the probability of losing in case auction is lost. 
    Thus, we can compute the log-likelihood function on partially observed data. Taking the likelihood for winning auctions, the corresponding negative log-likelihood for all the winning auctions is given by
    $\sum_{i \in \CW} -\log (\sum_{k=1}^{K} \frac{\pi_k(\x_i)}{\sigma_k(\x_i)} \phi(\frac{\w_i-\mu_k(\x_i)}{\sigma_k(\x_i)}))$
    where, $\phi$ is the pdf of $\N(0,1)$.
    For losing bids, we can similarly compute the probability of losing based on the lower bound,
    $\Pr(\W_i >\bb_i)  = \sum_{k=1}^{K} \pi_k(\x_i) \Phi(\frac{\mu_k(\x_i)-\bb_i}{\sigma_k(\x_i)})$
    
    Negative log-probability of all the losing auctions from the mixture density is,
    \begin{align}
    \sum_{i \in \LL} -\log (\sum_{k=1}^{K} \pi_k(\x_i) \Phi(\frac{\mu_k(\x_i)-\bb_i}{\sigma_k(\x_i)}))
    \label{eq:losemdn}
    \end{align}
    where, $\Phi$ represents the cdf of $\N(0,1)$.
    
    From Figure \ref{fig:km-estimate}, recall that the distribution is not unimodal and has multiple peaks. To address the multi-modality of the data we used a mixture of multiple densities. The embedded deep network in the MCNet ($\M$) is trained to learn the mean and standard deviation parameters of each of the constituents of the mixture model as well as the corresponding weights. 
    Combining all the auctions, we get the following optimization function for censored data,
    
    \begin{align}
    \M^{\ast} = \mbox{arg}\max_{\M}\sum_{i \in \LL} \log (\sum_{k=1}^{K} \pi_k(\x_i) \Phi(\frac{\mu_k(\x_i)-\bb_i}{\sigma_k(\x_i)})) \notag\\
    +\sum_{i \in \CW} \log (\sum_{k=1}^{K} \frac{\pi_k(\x_i)}{\sigma_k(\x_i)} \phi(\frac{\w_i-\mu_k(\x_i)}{\sigma_k(\x_i)})
    \label{eq:mdn}
    \end{align}
    
    where $\M$ is the neural network (parameters).
    
    
    \subsection{Optimization} It is easy to compute gradients of Eq. ~\ref{eq:censored}, ~\ref{eq:parm_censored}, ~\ref{eq:mdn} with respect to all the parameters. 
    We used Adam optimizer for stochastic gradient optimization \cite{kingma2014adam}. 
    \section{Experimental Results}
    In this section, we discuss experimental settings, evaluation measures, and results.
    
    \subsection{Experimental Settings}
    
    \paragraph{\bf Datasets:} We ran experiments on the publicly available iPinYou dataset {\cite{zhang2014real}} as well as on a proprietary dataset collected from Adobe Adcloud (a DSP). The iPinYou dataset contains censored winning price information. Further experimentation was done on a sampled week's data from Adobe Adcloud.
    iPinYou data is grouped into two subsets: session 2 (dates from 2013-06-06 to 2013-06-12), and session 3 (2013-10-19 to 2013-10-27). We experimented with the individual dates within sessions as well. For all the datasets, we allocated 60\% for training, 20\% for validation and rest 20\% for testing. We report the average as well as the standard deviation over five iterations.  Similar to previous research  on the iPinYou dataset, we remove fields that are not directly related to the winning price at the onset \cite{wu2015predicting,wang2016functional}. The fields used in our methods are UserAgent, Region, City, AdExchange, Domain, AdSlotId, SlotWidth, SlotHeight, SlotVisibility, SlotFormat, Usertag. Every categorical feature (e.g City), is one-hot encoded, whereas every numerical feature (e.g Ad height) is categorized into bins and subsequently represented as one-hot encoded vectors.  This way, each bid request is represented as a high-dimensional sparse vector. Table ~\ref{tab:ipin} shows the statistics of sessions in the iPinYou datasets. The number of samples and win rates for individual dates are mentioned in Table.~\ref{tab:dates}.
    
    \begin{table}
        \caption{Basic statistics of iPinYou Sessions}
        \centering
        \begin{tabular}{c|c|c|c}
            \hline 
            Session    & sample & feature   & win rate ($\%$)  \\ 
            \hline 
            2    & 53,289,330 & 40,664  & 22.87 \\ 
            3    & 10,566,743  & 25,020  & 29.64  \\ 
            \hline 
        \end{tabular}
        
        \label{tab:ipin}
    \end{table}

    \paragraph{\bf Evaluation Settings:}
    Evaluation on partially observed data is difficult when the winning price is unknown especially for point estimation.  In {\cite{wu2015predicting}}, the authors simulated new synthetic data from the original winning auctions.  
    While the added censoring allows validating point estimate, it does not use the whole data (or the true distribution). We evaluate the performance of predicting the winning price distribution rather than the point estimate itself. 
    Thus we use the whole data without generating simulated censoring behavior. This setting is similar to earlier work on the survival tree-based method where the authors evaluated predicting the distribution and used the original data  \cite{wang2016functional}.

    \paragraph{\bf Parametric methods:}
    We compared the Censored Regression (CR) approach with our methods: Fully parametric Censored Regression (P-CR) and Mixture Density Censored Network (MCNet).
    For every method, we added an L2 regularization term for each weight vector to prevent over-fitting. For MCNet, we added an additional hyper-parameter on the number of mixtures. We chose a fully connected hidden layer with 64 nodes with RelU activation function as the architecture. Our framework is general and can be extended to multiple layers. The number of mixture components was varied from 2-4 for individual dates and 2-8 for the experiments on the two sessions. We used Adam optimizer with a learning rate of $10^{-3}$. Mini-batch training was employed due to the high volume of the data and we fixed the batch size to 1024 samples. We employed early stopping on the training loss and do not observe the validation loss for early stopping. This way, all the methods are treated similarly. The L2 regularization was varied from $10^{-6} \mbox{ to } 10$ (in log scale). We implemented the parametric methods in Tensorflow \cite{abadi2016tensorflow}. For the initialization of weight vectors, we sampled randomly from the standard normal distribution in all our experiments.
    
    Recently extending Censored Regression (CR), in {\cite{wu2018deep}}, the authors proposed to use deep model (DeepCR) to parameterize the mean to provide more flexibility in the point estimation. Additionally, the authors proposed to use Gumbel distribution for point estimation. Note that, MCNet generalizes the DeepCR model when using only one mixture component and Gumbel as the base distribution. We did not see much improvement when using Gumbel to parameterize mixture components with our initial experiments. With enough Gaussian mixture components, MCNet can approximate any smooth distribution. As neural architecture is not the primary motivation for this paper, we do not discuss different architectures or distributions in this paper.
    
    \paragraph{\bf Non-parametric methods:}
    To the best of our knowledge, parametric methods and non-parametric methods were not compared together for winning price distribution estimation in earlier research. We compared our approaches with non-parametric approaches based on Kaplan-Meier (KM) estimate and the Survival tree (ST) method. The KM and ST based methods produce winning price distributions until the maximum bid price since the winning price behavior above that is unknown. To represent a complete landscape with the probability distribution summing to one, we introduce an extra variable representing the event that the winning price is beyond the maximum bid price. For the Survival tree, we varied the tree height from 1-20. 
    
    In the ST method, the Survival tree is built by running an Expectation Maximization (EM) algorithm for each field to cluster similar attributes. If data has $F$ fields and the average number of attributes in each field is $K$, then for $n$ samples, the EM  algorithm takes, $\OO(FKln)$ steps to cluster features based on their density for $l$ iterations. With depth $d$, total complexity becomes $\OO(FKlnd)$. Given this runtime, we could not run ST using all attributes of {\it Domain}, {\it SlotId} fields (these fields were removed in previous research \cite{wang2016functional}). We trimmed the {\it Domain} and {\it SlotId} features by combining the attributes that appeared less than $10^3$ times. We created {\it ``other domains"} and {\it ``other slot ids"} bins for these less frequent attributes. This improved the time complexity and made the method viable.
    But for the CR-based methods, we could easily relax this threshold and trimmed both the features where the attributes appeared less than 10 times in the dataset in either session. For a fair comparison, whenever we use the same feature trimming in the parametric methods as ST, we denote using $\mbox{CR}^{\ast}, \mbox{P-CR}^{\ast},\mbox{MCNet}^{\ast}$. Note that parametric methods can scale easily whereas the non-parametric ST method cannot. 
    \begin{figure*}%
        \centering
        \subfloat[Session 2 Performance]{{\includegraphics[width=6.1cm]{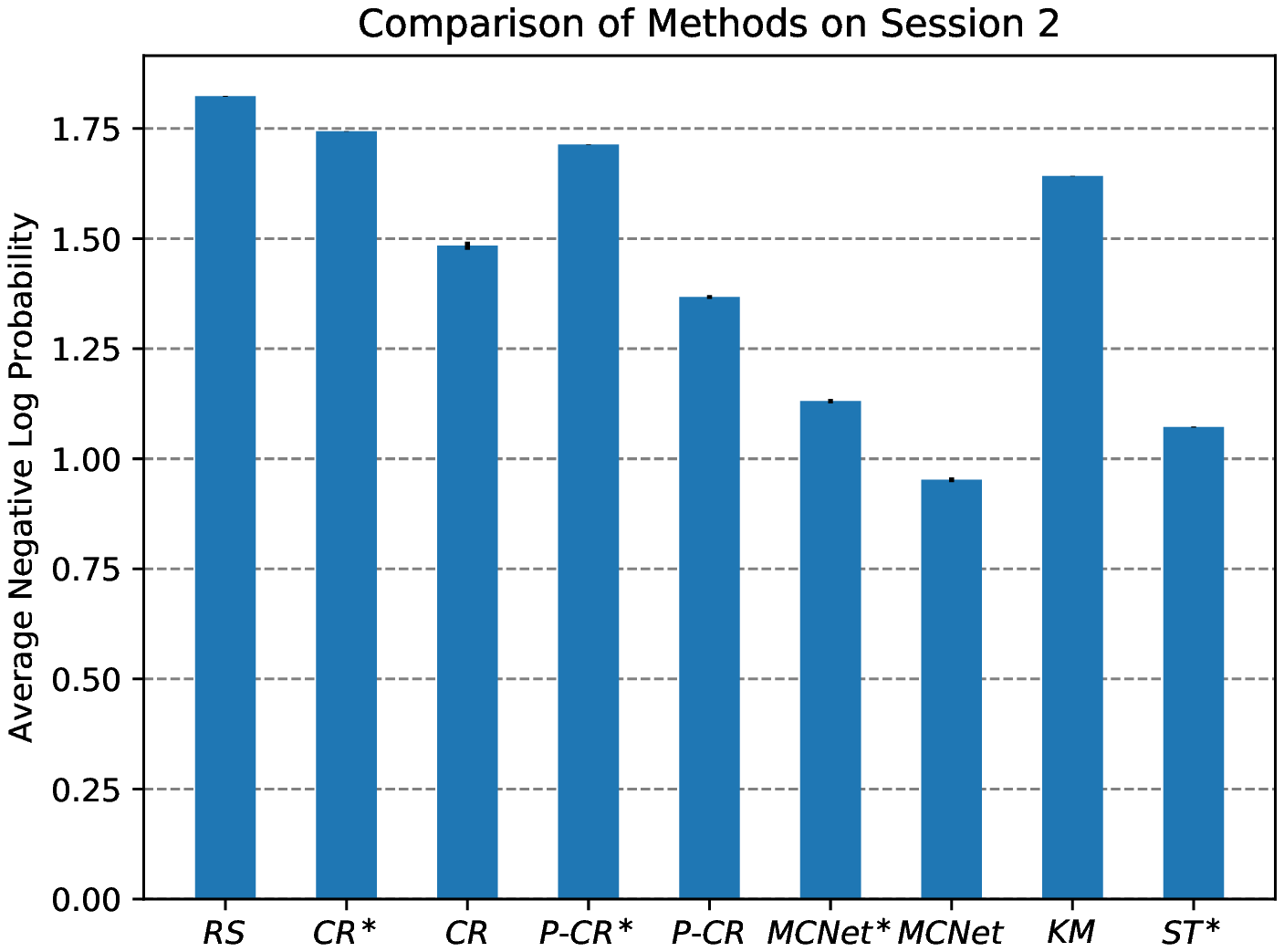} }}%
        \subfloat[Session 3 Performance]{{\includegraphics[width=6.1cm]{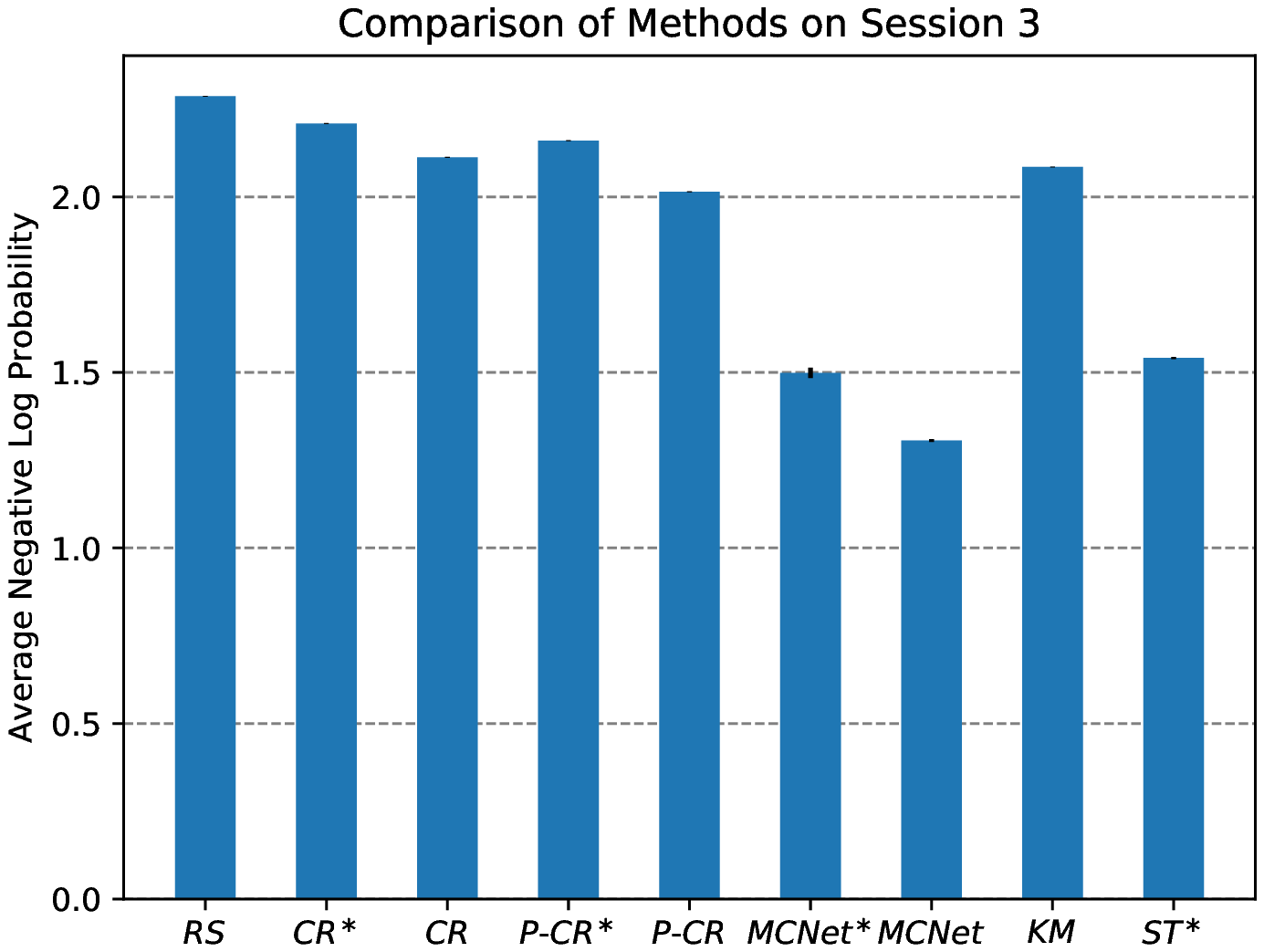} }}%
        \caption{iPinYou session's \ANLP. Error bar represents the standard deviation.}
        \label{fig:session}%
    \end{figure*}

    \paragraph{\bf Baseline Method:}
    We also propose a simple baseline method and compare it with other methods.
    The baseline algorithm picks a winning price randomly conditioned on the win rate. We denote this as the Random Strategy (RS). Formally, let the maximum bid price be $z$ and probability of a win be $p$. Then, the probability that the winning price is $w$ is given by
    \begin{align*}
    P(\W =w) &= \frac{p}{z} \mbox{ if } w\in[0,z], \mbox{ and }  0 \mbox{ if } w <0\ 
    \mbox{and }&\int_z^{\infty} \Pr(\W =w)dw  = 1-p
    \end{align*}
    Thus with probability $1-p$, it predicts the event that winning price is greater than max bid price and with probability $p$ it draws from $\UU(0,z)$ where $\UU$ is the Uniform distribution.
    
    \subsection{Evaluation Measure}
    Our objective is to learn the distribution of the winning price, rather than the point estimate. Hence, we choose Average Negative Log Probability (\ANLP) as our evaluation measure similar to \cite{wang2016functional}. \ANLP is defined as,
    \begin{align*}
    \mbox{\ANLP}= &-\frac{1}{N}\Big(\sum_{i\in \CW}\log \Pr(\W_i =\w_i)+
    \sum_{i\in \LL} \log \Pr(\W_i \geq \bb_i)\Big)
    \end{align*}
    where $\CW$ represents the set of winning auctions, $\w_i$ represents winning price of the $i^{\mbox{th}}$ winning auction, $\LL$ is the set of losing auctions, $\bb_i$ is the bidding price for the $i^{\mbox{th}}$ losing auction, and $|\CW| + |\LL| = N$.

    Note that, we computed pdf for winning auctions and probability (or the CDF) for losing auctions while optimizing. While the CDF represents the probability of the event, density does not represent probability. Additionally, bid prices are an integer. The KM method estimates the probability on those discrete points. However, parametric approaches estimate a continuous random variable whose probability at any discrete point is $0$. To treat losing bids and winning bids similarly in evaluation, we use quantization trick on the continuous random variable \cite{gersho2012vector}.
    For the parametric approaches, the estimate $\W_i$ is a continuous random variable. We discretized the random variable $\W_i$ as follows,
    $\W_i^{\mbox{bin}}  = l, \mbox{ if } \W_i \in (l-0.5,l+0.5]$ where $l$ is an integer.
    Thus, for winning auctions $\CW$ with winning price $\w_i$, quantized probability is, 
    \begin{align*}
    \Pr(\W_i^{\mbox{bin}} =\w_i) & =\Pr(\W_i\leq \w_i +0.5) - \Pr(\W_i\leq \w_i-0.5)    
    \end{align*}
    For losing auctions $\LL$, the quantized probability is,
    $\Pr(\W_i^{\mbox{bin}} \geq \bb_i)  =\Pr(\W_i\geq \bb_i-0.5)$.
    Using quantization technique, winning bids and losing bids are treated similarly for all methods. 
    
    \subsection{Experimental Results}
    \begin{figure}%
        \centering
        \subfloat[Varying Height of ST]{{\includegraphics[width=6.1cm]{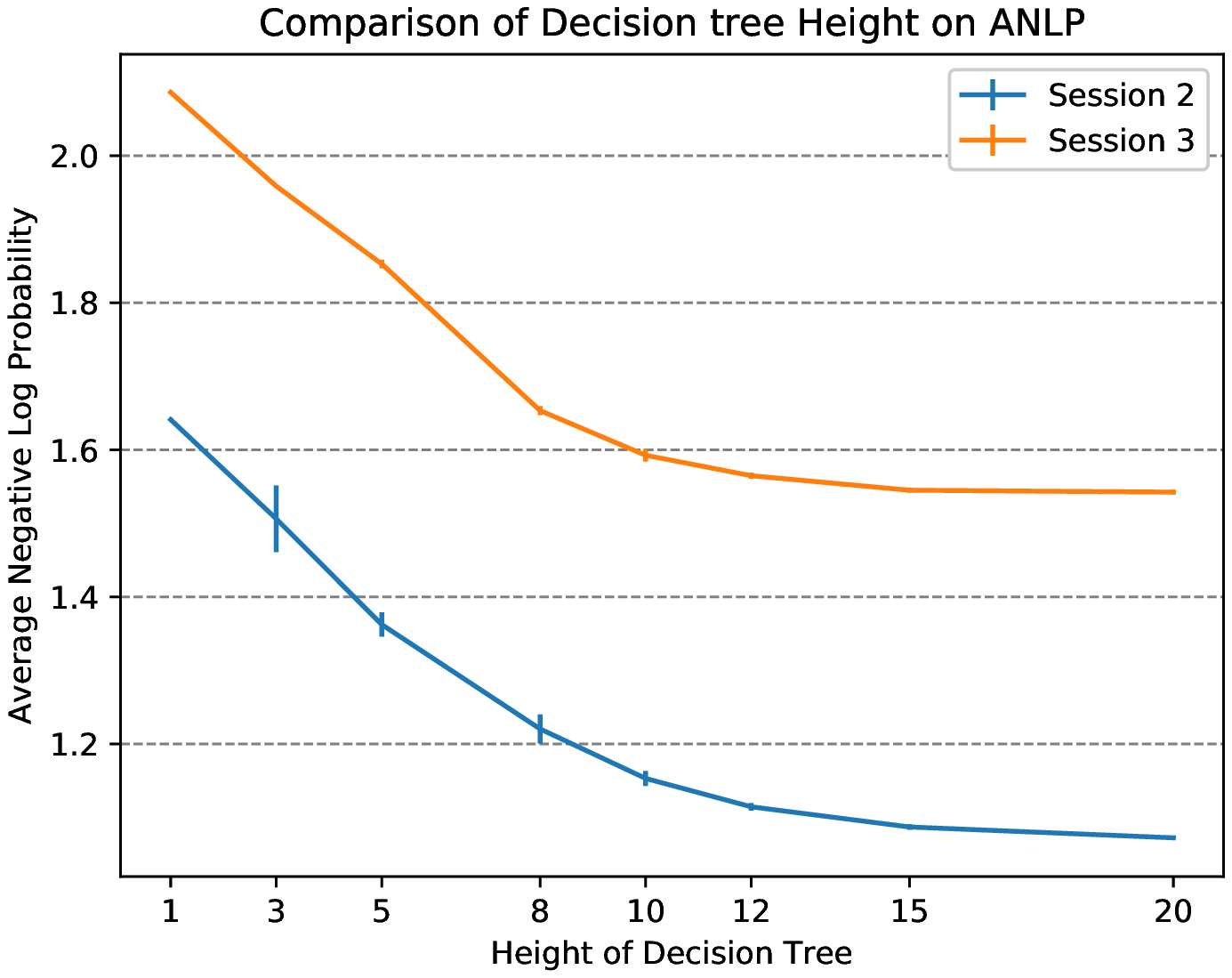} }}%
        \subfloat[Varying mixture components]{{\includegraphics[width=6.1cm]{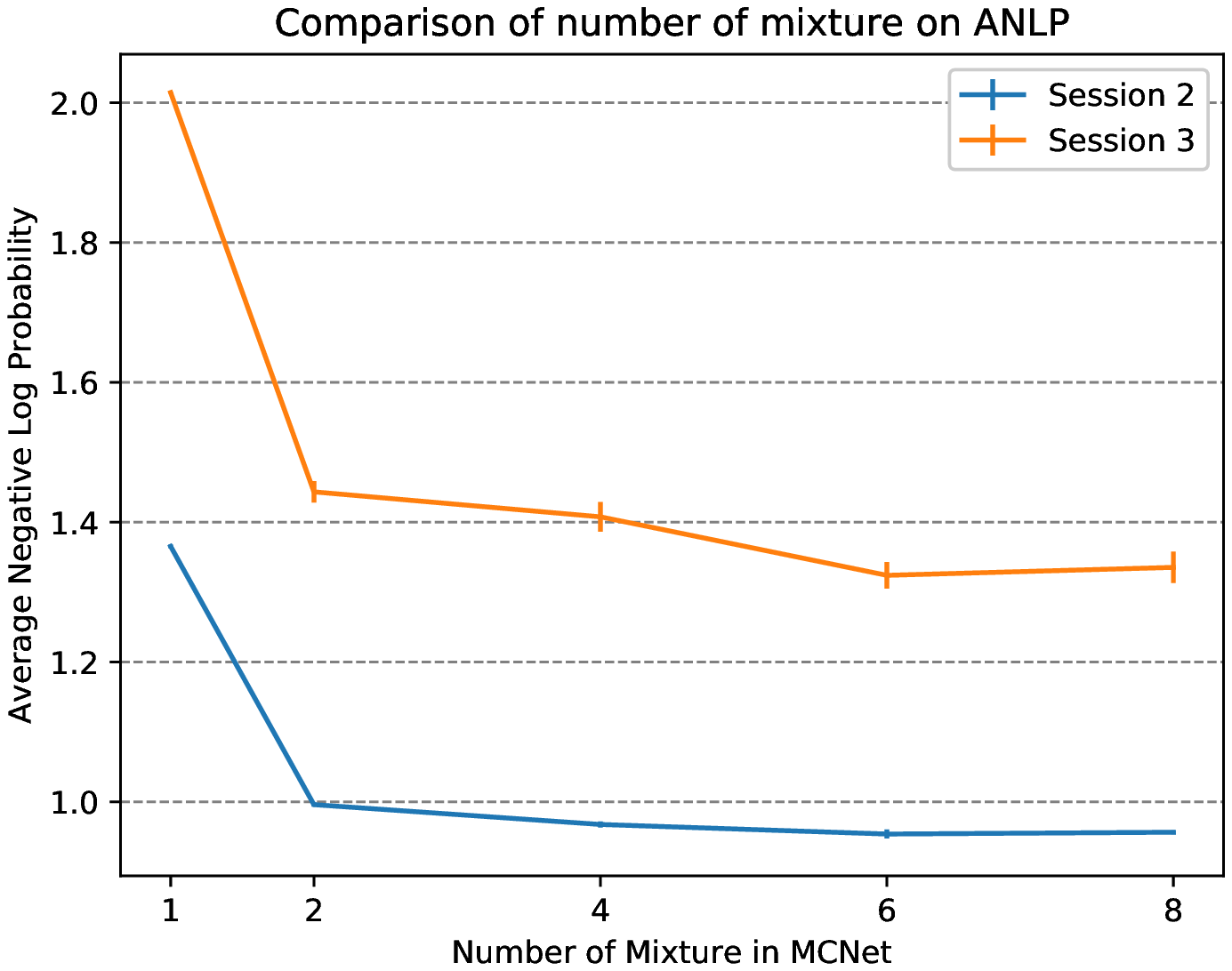} }}%
        \caption{Hyper-parameter effect on \ANLP}
        \label{fig:component}%
    \end{figure}
    \begin{table*}[t]
        \centering
        \caption{\ANLP on Session 2 and 3  individual dates. We report std only if it is higher than $0.01$}
        \resizebox{\columnwidth}{!}{%
            \begin{tabular}{c|c|c|c|c|c|c|c| c}
                \hline 
                Date & $\approx$n($\times10^6$) & wr($\%$) & RS     & CR  & p-CR & MCNet & KM  & ST \\ 
                \hline 
                2013-06-06&9.58& 18.93    & 1.55&$1.212$ & $1.081$  & $\mathbf{0.756}$ & $1.403$ &$0.956$  \\
                2013-06-07&11.13& 16.16&1.35    & $1.036$  & $0.913$  & $\mathbf{0.641}$ $\pm$ $0.02
                $ & $1.211$  & $0.823$ \\ 
                2013-06-08&5.22&31.17 &     2.37& $1.946$ & $1.695$  & $\mathbf{1.311}$ $\pm$ $0.02
                $ & $2.131$  & $1.527$ \\
                2013-06-09&11.88 &13.85 &1.17    & $0.887$ & $0.784$  & $\mathbf{0.574}$ $\pm$ $0.03
                $ & $1.071$ & $0.710$\\
                2013-06-10 &5.61& 34.06 & 2.55    & $2.130$ & $1.809$ & $\mathbf{1.252}$ $\pm$ $0.06
                $ & $2.234$ &$1.502$  \\
                2013-06-11 &5.09& 34.13 & 2.56    &$2.128$  & $1.810$  & $\mathbf{1.351}$ 
                & $2.248$  & $1.552$ \\
                2013-06-12 &4.75& 34.68 &    2.59&$2.189$  & $1.914$ & $\mathbf{1.364}$ $\pm$ $0.04$ & $2.273$  & $1.572$ \\ 
                \hline 
                \\
                \hline
                2013-10-19&0.35& 64.58 & 4.31 &$4.135$ & $4.285$ $\pm$ $0.08$ & $\mathbf{2.791}$ $\pm$ $0.05
                $ & $3.659$  & $3.056$ $\pm$ $0.02$ \\
                2013-10-20&0.32& 65.48 &4.33    &  $4.167$  & $4.287$ $\pm$ $0.15$ & $\mathbf{2.768}$ $\pm$ $0.1
                $ & $3.737$ & $3.159$ \\
                2013-10-21 & 1.54&  54.59&    3.77&$3.466$ & $3.515$  & $\mathbf{2.338}$ $\pm$ $0.03
                $ & $3.272$ & $2.529$  \\
                2013-10-22&1.21& 56.00 &3.85    & $3.641$  & $3.569$ & $\mathbf{2.576}$ $\pm$ $0.03
                $ & $3.428$ & $2.779$ \\
                2013-10-23 &1.57&14.30 &1.22    & $1.060$& $1.033$ & $\mathbf{0.854}$ $\pm$ $0.02
                $ & $1.157$ & $0.963$\\
                2013-10-24 &2.18&  11.23     &0.985& $0.831$  & $0.824$ & $\mathbf{0.618}$ & $0.904$  &$0.698$ $\pm$ $0.01$ \\
                2013-10-25 &2.23& 14.23     &1.21&$1.015$ & $0.998$ & $\mathbf{0.771}$ $\pm$ $0.03
                $ & $1.131$ & $0.888$ \\
                2013-10-26&0.53&49.90    &3.51& $3.432$ & $3.433$ $\pm$ $0.01$ & $\mathbf{2.577}$ $\pm$ $0.09
                $ & $3.228$ & $2.931$ $\pm$ $0.01$ \\
                2013-10-27&0.59&18.45 & 1.53    &  $1.367$  & $1.361$  & $\mathbf{0.937}$ $\pm$ $0.03
                $ & $1.348$  & $1.104$ $\pm$ $0.02$\\
                \hline 
            \end{tabular}
        }
        
        \label{tab:dates} 
    \end{table*}
    
    \begin{table}
        \centering
        \caption{\ANLP on Adobe AdCloud Dataset}
        \resizebox{\columnwidth}{!}{%
            \begin{tabular}{c|c | c | c|c|c}
                \hline 
                & CR & P-CR & MCNet & KM & ST  \\ 
                \hline 
                \ANLP  &$0.4744$ $\pm$ $0.01$  & $0.4722$ $\pm$ $0.01$ & $\mathbf{0.3477}$ $\pm$ $0.01$ & $0.4671$ $\pm$ $0.01$  & $0.4213$ $\pm$ $0.02$ \\ 
                \hline         
        \end{tabular} }
        
        \label{tab:adcloud}
    \end{table}
    In this section, we discuss quantitative results on iPinYou sessions 2 and 3. In Table \ref{tab:dates}, we provide average \ANLP over different dates as well as  the standard deviation (std) numbers.  In figure ~\ref{fig:session}, we mention the result on each session as a whole. Moreover, we plot how number of mixture components as well as tree depth affect the result for MCNet and ST respectively in Figure \ref{fig:component}.
    In sessions 2 and 3 where we include all the dates, we also added the ST method for comparison. As ST did not run with large feature space, we also added  $\mbox{CR}^{\ast}, \mbox{P-CR}^{\ast},\mbox{MCNet}^{\ast}$ for parity (where number of feature was small for all methods). 
    
    From Table~\ref{tab:dates}, it is evident, P-CR improves upon CR on most dates (except with low volume dates) asserting the violation of fixed standard deviation assumption. While for P-CR, improvement is around $5\%$-$10\%$, MCNet improves CR by more than $30\%$ on all dates.  Improvement of MCNet re-verify our assumption about the multi-modal nature of the winning price distribution. CR performs better than both RS as well as KM. This is expected as the non-parametric KM estimate does not use any features. However, KM improves RS by around {$10\%$} on all dates. ST improves CR and P-CR significantly implying the significance of non-parametric estimators.
    
    In Figure ~\ref{fig:session}, one can see similar trends over CR, P-CR, and MCNet. With feature trimming, $\mbox{MCNet}^{\ast}$ performs similarly to ST methods. This is expected as both ST and MCNet can predict arbitrary smooth distributions. 
    Although, when the MCNet approach is restricted to fewer features ($\mbox{MCNet}^{\ast}$) on the average it performs similarly to ST, the benefits of parametric methods come from the fact that parametric approaches are scalable to large feature as well as input space. It may be observed that the performance of MCNet improves ST by more than $10\%$ on both sessions. While we used only one hidden layer for MCNet, any deep network can be used to parameterize the mixture density network for potentially improving the MCNet results even further.
    
    In Figure~\ref{fig:component}(a), we plot \ANLP for different depths of the decision trees. It can be observed that for ST, the performance saturates around depth 15. In Figure ~\ref{fig:component}(b), we also show how the varying number of mixture components impacts \ANLP.  On the larger dataset of Session 2, \ANLP stabilizes to a low value at 4 mixture components. However, for session 3, \ANLP starts increasing beyond 6 mixture components, implying over-fitting.
    
    \paragraph{\bf  Results on Adobe AdCloud Dataset: }
    We also tested our methods on  Adobe Advertising Cloud (DSP) offline dataset.  We collected a fraction of logs for one week. The number of samples was  $31,772,122$ and the number of features was $33,492$.  It had similar categorical as well as real-valued features. We use the same featurization framework and represented each bid request with a sparse vector.
    In Table ~\ref{tab:adcloud}, we report the \ANLP results, using the same experimental setup. 
    Note that, MCNet improves CR by $25\%$ while it improves ST by more than $10\%$. In this dataset, we do see only marginal improvement over using P-CR. 
    
    \section{Discussion \& Future Work} In this paper,
    we particularly focus on one of the central problems in RTB, the winning price distribution estimation. In practice, DSP depends on the estimated bid landscape to optimize it's bidding strategy. From a revenue perspective, an accurate bid landscape is of utmost importance. While, non-parametric methods can estimate arbitrary distributions, in practice, it is challenging to scale on large datasets. On the other hand, widely used parametric methods, such as Censored Regression in its original form is highly restrictive. We proposed a novel method based on Mixture Density Networks to form a generic framework for estimating arbitrary distribution under censored data. MCNet generalizes a fully parametric Censored regression approach with the number of mixture components set to one. Additionally, Censored regression is a special case of fully parametric censored regression where the standard deviation is fixed. We provided extensive empirical evidence on public datasets and data from a leading DSP to prove the efficacy of our methods. 
    While the mixture of (enough) Gaussian densities can approximate any smooth distribution, further study is needed on the choice of base distribution.
    A more subtle point arises when learning with censored data as we do not observe any winning price beyond the maximum bidding price. Without any assumptions on the distribution, it is not provably possible to predict the behavior in the censored region. Non-parametric methods only learn density within the limit of maximum bidding price while under strong assumptions of standard distributions, censored regression predicts the behavior of winning price in the censored region. Although MCNet can approximate any smooth distribution, beyond the maximum bidding price, it leads to non-identifiability similar to the KM estimate.  
    It would be interesting to explore combining MCNet with distributional assumptions where the winning price is never observed.
    \balance
    

\end{document}